\documentclass[letterpaper]{article} % DO NOT CHANGE THIS
\usepackage{aaai2026}  % DO NOT CHANGE THIS
\usepackage{times}  % DO NOT CHANGE THIS
\usepackage{helvet}  % DO NOT CHANGE THIS
\usepackage{courier}  % DO NOT CHANGE THIS
\usepackage[hyphens]{url}  % DO NOT CHANGE THIS
\usepackage{graphicx} % DO NOT CHANGE THIS
\usepackage{natbib}  % DO NOT CHANGE THIS AND DO NOT ADD ANY OPTIONS TO IT
\usepackage{caption} % DO NOT CHANGE THIS AND DO NOT ADD ANY OPTIONS TO IT
\usepackage{dsfont}
\usepackage{hyperref}

\usepackage{algorithm}
\usepackage{algorithmic}
\usepackage{amsmath}
\usepackage{amssymb}
\usepackage{amsfonts}
\usepackage{cleveref}
\usepackage{booktabs}
\usepackage{graphicx}        % for \resizebox
\usepackage[table]{xcolor}   % for row colors
\usepackage{multirow} 
\usepackage{makecell}
\usepackage{pifont}
\newcommand{\cmark}{\ding{51}} % 对号
\newcommand{\xmark}{\ding{55}} % 叉号

\usepackage{newfloat}
\usepackage{listings}
\DeclareCaptionStyle{ruled}{labelfont=normalfont,labelsep=colon,strut=off} % DO NOT CHANGE THIS
\floatstyle{ruled}
\newfloat{listing}{tb}{lst}{}
\floatname{listing}{Listing}
\title{vMFCoOp: Towards Equilibrium on a Unified Hyperspherical\\  Manifold for Prompting Biomedical VLMs}
\author{
   Minye Shao\textsuperscript{\rm 1},
   Sihan Guo\textsuperscript{\rm 1},
    Xinrun Li\textsuperscript{\rm 1},
    Xingyu Miao\textsuperscript{\rm 1},
    Haoran Duan\textsuperscript{\rm 2},
    Yang Long \textsuperscript{\rm 1}\thanks{Corresponding author.}
}
\affiliations{
    \textsuperscript{\rm 1}Department of Computer Science, Durham University, Durham, UK\\
    \textsuperscript{\rm 2}Department of Automation, Tsinghua University, Beijing, China\\
     \{minye.shao, yang.long\}@ieee.org\\
    \{sihan.guo, xinrun.li, xingyu.miao\}@durham.ac.uk\\
    haoranduan@mail.tsinghua.edu.cn\\

}

\usepackage{bibentry}
\begin{document}

\maketitle

\begin{abstract}
Recent advances in context optimization (CoOp) guided by large language model (LLM)–distilled medical semantic priors offer a scalable alternative to manual prompt engineering and full fine-tuning for adapting biomedical CLIP-based vision-language models (VLMs). However, prompt learning in this context is challenged by semantic misalignment between LLMs and CLIP variants due to divergent training corpora and model architectures; it further lacks scalability across continuously evolving families of foundation models. More critically, pairwise multimodal alignment via conventional Euclidean-space optimization lacks the capacity to model unified representations or apply localized geometric constraints, which tends to amplify modality gaps in complex biomedical imaging and destabilize few-shot adaptation. In this work, we propose vMFCoOp, a framework that inversely estimates von Mises–Fisher (vMF) distributions on a shared Hyperspherical Manifold, aligning semantic biases between arbitrary LLMs and CLIP backbones via Unified Semantic Anchors to achieve robust biomedical prompting and superior few-shot classification. Grounded in three complementary constraints, vMFCoOp demonstrates consistent improvements across 14 medical datasets, 12 medical imaging modalities, and 13 anatomical regions, outperforming state-of-the-art methods in accuracy, generalization, and clinical applicability. \textit{This work aims to continuously expand to encompass more downstream applications, and the corresponding resources are intended to be shared through} {\hypersetup{urlcolor=black}\href{https://github.com/VinyehShaw/UniEqui}{https://github.com/VinyehShaw/UniEqui}}.
\end{abstract}

% Uncomment the following to link to your code, datasets, an extended version or similar.
% You must keep this block between (not within) the abstract and the main body of the paper.

\section{Introduction}

Recent advances in vision-language models (VLMs), such as CLIP \cite{radford2021learning}, have significantly advanced general vision tasks by aligning visual and linguistic semantics through large-scale contrastive learning, enabling strong zero-shot and few-shot generalization. Biomedical images exhibit highly structured semantics and demand domain-specific knowledge, making it challenging for VLMs pretrained on natural images to generalize effectively. Previous adaptations of CLIP to the medical domain often require heavy supervision, limiting scalability and generalization under low-data settings.
In response, recent medical-specific CLIP variants \cite{wang2022medclip,eslami2023pubmedclip} have emerged, pretrained on large-scale medical image–text pairs, and exhibit improved performance \cite{zhang2022contrastive}.   However, these VLMs largely hinge on the quality of the textual prompts used, and full fine-tuning of large-scale pretrained models is often impractical due to computational and data constraints \cite{schmalfuss2025parc}. To address this, CLIP-based prompt learning has emerged as a lightweight adaptation strategy that optimizes input prompts to guide cross-modal predictions without modifying model parameters. Methods such as CoOp \cite{zhou2022learning} learn tunable context vectors to construct task-specific semantic prompts, substantially enhancing CLIP’s performance across diverse downstream tasks.

Unlike natural images, biomedical images present greater visual complexity and modality diversity, with fine-grained structures, strong anatomical priors, and significant cross-scale variations as illustrated by the medical imaging cases in \Cref{ov}, imposing higher demands on model discrimination. Meanwhile, the acquisition of high-quality labeled data is severely limited by privacy concerns, high annotation costs, and the rarity or inaccessibility of certain patient populations, making few-shot learning a central paradigm in medical imaging \cite{huang2023rethinking}. Multimodal large language models (MLLMs), trained on rich medical corpora, show promising capabilities in basic recognition tasks but remain unstable in structural understanding and high-level semantics, often exhibiting hallucinations, particularly in complex scenarios like multi-stage diagnosis or surgical planning. In contrast, CLIP-based medical VLMs offer lightweight, interpretable architectures with stronger visual–language alignment and better clinical deployability \cite{chen2025mimo}. To leverage the strengths of both LLMs and CLIP, recent work such as BiomedCoOp \cite{koleilat2025biomedcoop} introduces LLM-generated prompts to guide CLIP’s contextual learning. It implicitly assumes a direct compatibility between the language priors of LLMs and the multimodal embedding space of CLIP. However, prompt learning across such heterogeneous modalities entails more than simple token-level substitution; it requires reconciling differing semantic abstractions, representational granularities, and alignment dynamics. As with many conventional approaches to cross-modal alignment, it often relies on independent pairwise matching within flat Euclidean spaces, which is insufficient to capture the intrinsic relational geometry and directional semantics among modalities \cite{zheng2025hierarchical}.

To overcome the limitations of flat Euclidean alignment in cross-modal prompt learning, recent efforts have explored manifold optimization to better respect the geometric structure of high-dimensional representation spaces \cite{park2024prompt}. In particular, methods combining CoOp-style prompt tuning with geometry-aware learning have demonstrated improved robustness and adaptability by constraining features or prompts to lie on Riemannian manifolds such as hyperspheres \cite{cho2023distribution}. However, these approaches often treat vision and language representations independently, neglecting the heterogeneous semantic granularity and alignment dynamics between LLMs and CLIP. Moreover, existing designs typically impose fixed geometric priors without accounting for distributional mismatch or modality-specific biases, leading to unstable convergence or suboptimal generalization under few-shot regimes \cite{wang2023few}. To address all the above challenges, we introduce vMFCoOp, a novel biomedical prompting framework that leverages a unified hyperspherical manifold and inverse vMF mapping estimation to achieve stable, generalizable, and model-agnostic few-shot learning. In summary, our contributions are three-fold:
\begin{itemize}
 \item This is the first framework to introduce an inverse-estimation based on hyperspherical manifold probabilistic modeling, enabling explicit capture and reconciliation of semantic biases inherent in diverse foundation models, and ensuring scalable adaptability across rapidly evolving model families.
    \item  We revisit few-shot prompt learning for biomedical VLMs on a Unified Hyperspherical Manifold, where semantic anchors bridge cross-modal gaps and improve generalization beyond Euclidean-space approaches.

    \item Extensive validation on 14 realistic clinical few-shot scenarios, including challenging datasets from UK Biobank, confirms that vMFCoOp substantially improves biomedical prompt-based classification performance, supporting its potential for clinical deployment.
\end{itemize}

\section{Related Work}
\subsection{Few-shot Adaptation of VLMs}
Prompt-based methods have emerged as efficient solutions for adapting vision-language models (VLMs) like CLIP \cite{radford2021learning} to downstream tasks with limited data. CoOp \cite{zhou2022learning} learns continuous prompts but overfits to base classes; CoCoOp \cite{zhou2022conditional}  mitigates this via instance-conditioned prompts for better generalization. ProGrad \cite{zhu2023prompt} and KgCoOp \cite{yao2023visual} introduce gradient or knowledge alignment to retain pretrained semantics, yet trade off flexibility or expressiveness. PLOT \cite{chen2022prompt} enhances fine-grained alignment using multi-prompt optimal transport, while MaPLe \cite{khattak2023maple} and PromptSRC \cite{khattak2023self} improve prompt diversity and generalization. Adapter-based approaches like CLIP-Adapter \cite{gao2024clip} and Tip-Adapter \cite{zhang2022tip} offer efficient black-box tuning, but struggle with hyperparameter sensitivity and underutilized text features. In the medical domain, MedFocusCLIP \cite{arora2025medfocusclip} argues that CoOp is insufficient for fine-grained medical image classification and introduces SAM-based ROIs to guide attention toward discriminative regions; CLIPath \cite{lai2023clipath} introduces residual adapters for pathology. Despite progress, existing methods often lack unified representations, explicit semantic alignment, or stable adaptation across diverse model families. 

\subsection{Prompt Learning with Manifold}
To improve prompt tuning's adaptability and generalization, manifold-based extensions to CoOp have been explored. ProMetaR \cite{park2024prompt} frames prompt tuning as bi-level meta-regularization with Manifold Mixup to enhance generalization, but depends on curated meta-tasks and lacks scalability to diverse models and evolving prompts. ProDA \cite{lu2022prompt} models prompts as Gaussians to capture intra-class variance, yet assumes unimodal linear geometry. ProGrad \cite{zhu2023prompt} constrains updates along zero-shot gradient directions on the CLIP manifold, improving stability but limiting semantic flexibility. \cite{chen2024hyperbolic} project prompts into a Poincaré disk to model textual hierarchies, but ignore cross-modal alignment, vision-language gaps, and VLM scalability. MERU \cite{desai2023hyperbolic} uses hyperbolic spaces for semantic hierarchies and uncertainty, improving interpretability and alignment but requiring complex optimization and curvature tuning. DualCoOp \cite{sun2022dualcoop} embeds prompts in label-conditioned manifolds for multi-label tuning, yet overlooks inter-label dependency and efficiency. Collectively, these methods show that distributional or non-Euclidean manifolds enhance prompt alignment and generalization beyond Euclidean tuning, though scalability and adaptability remain key challenges in biomedical settings.

\begin{figure*}
    \centering
   
    \includegraphics[width=\textwidth]{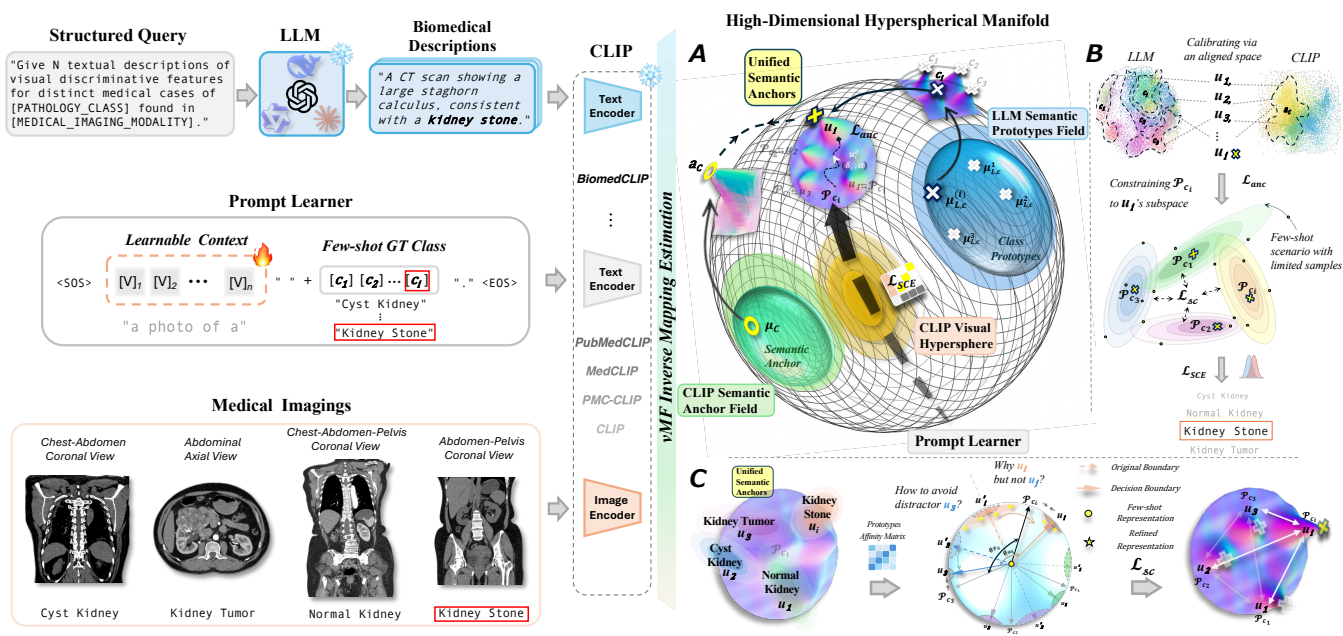}

   \vspace{20pt}
  \caption{vMFCoOp framework combines structured prompts from an arbitrary LLM, learnable context tokens, and any biomedical CLIP variant to model complex medical images under few-shot settings, with vMF inverse mapping estimation enabling unified semantic alignment on the manifold, guided by 3 constraints. \textbf{A}.Unified semantic anchor construction and prompt optimization on the hyperspherical manifold. \textbf{B}.Low-dimensional overview of the aligned framework for intuitive understanding: vMFCoOp seeks equilibrium on a unified hyperspherical manifold by calibrating priors from heterogeneous, evolving VLM/LLM families through estimation and anchoring that stabilizes directional semantics, and unified optimization within the non-Euclidean hyperspherical space, thereby reconciling cross-model semantic biases, accommodating fine-grained biomedical variability, and stabilizing few-shot clinical adaptation. \textbf{C}.Details of the $\mathcal{L}_\mathit{\boldsymbol{sc}}$ loss: Decision boundaries evolve from initial equal-angle partitions to large-margin separations as the temperature $\tau$ is annealed. 
Within-class representations are refined from broad angular spread to compact clusters. 
For visualization, vectors are projected onto their $S^1$ subspace (angles preserved), while optimization is performed on the full hypersphere $S^{d-1}$.}
   \label{ov}
     
\end{figure*}

\subsection{vMF-based Representation Learning}
The von Mises–Fisher (vMF) distribution models data on the unit hypersphere using a mean direction and concentration, serving as a fundamental distribution for statistical analysis on Riemannian manifolds and offering promising potential for modern representation learning \cite{mardia2009directional}. ProCo \cite{du2024probabilistic} models class-wise features with vMF distributions to derive closed-form contrastive losses without large batches, achieving strong long-tailed recognition but lacking cross-modal semantic alignment. T-vMF \cite{kobayashi2021t} introduces a vMF-based angular similarity to compact feature clusters for visual classification on the hypersphere, but lacks extension to cross-modal alignment. In multimodal contexts, CLIP-Enhance \cite{sandhuclip} fits vMF mixtures to CLIP embeddings to address modality gaps and long-tail bias, but relies on extra clustering. SSP \cite{zhu2024selective} identifies vMF-based misalignment in CLIP and proposes a projection-based modality bridging strategy, though without class-level semantic adaptability. Meanwhile, vMF Loss \cite{scott2021mises} leverages spherical embedding geometry for calibrated classification, but lacks support for prompt-based or multimodal tasks. Our task naturally resides on a hyperspherical manifold, as both CLIP and LLM embeddings are normalized and lie on the unit sphere, making vMF modeling well-suited for multimodal representation alignment with geometric consistency.

\section{Methodology}

Our vMFCoOp framework flexibly plugs into any LLM and biomedical-CLIP backbone to jointly learn prompt contexts. It imposes complementary constraints on a shared hyperspherical manifold, guiding prompt and visual embeddings toward a unified semantic anchor field reconciling multimodal textual biases via the \textit{vMF Inverse Mapping Estimation}, in turn mitigating textual semantic transfer misalignment across diverse model priors and granularity levels in biomedical imaging. \Cref{ov}B provides an intuitive lower-dimensional overview of the hyperspherical manifold, highlighting three hierarchical constraints: \textit{Semantic Anchor Loss} ($\mathcal{L}_\mathit{\boldsymbol{anc}}$) encourages the prompt context embeddings to converge toward a unified semantic anchor by calibrating the inherently mismatched LLM and CLIP embedding spaces into an aligned subspace, \textit{Spherical Contrastive Loss} ($\mathcal{L}_\mathit{\boldsymbol{sc}}$)  introduces supervised angular margins over the unit hypersphere to enhance inter-class separability within the shared manifold, and \textit{Symmetric Cross-Entropy Loss} ($\mathcal{L}_\mathit{\boldsymbol{SCE}}$) jointly minimizes forward and reverse divergences between image- and prompt-based class distributions. Consequently, few-shot samples converge toward multimodal semantic equilibrium via this unified anchor-guided manifold space to facilitate class-level prompt prediction. More mathematical derivations and additional theoretical proofs are provided in the \textit{Proofs document}.

\subsection{Pre-trained Backbones and Prompt Learner Setup}
We use BiomedCLIP and other biomedical CLIP variants take a tokenized prompt sequence \(\boldsymbol{X_t}\in \mathbb{R}^{B \times L} \) and a preprocessed RGB medical image \(\boldsymbol{X_v}\in \mathbb{R}^{B \times 3 \times H \times W} \) as input, and encodes them into the embedding space via a text encoder \(\boldsymbol{E_t}\) and a vision encoder \(\boldsymbol{E_v}\), respectively. The text \( \boldsymbol{T}\) and vision embeddings \(\boldsymbol{V}\) are given by \(\boldsymbol{E_t(X_t)} \in \mathbb{R}^{C\times D} \) and \( \boldsymbol{E_v(X_v)} \in \mathbb{R}^{B \times D} \), where \( B \) is the batch size, \( L \) is the maximum token length, \( H \) and \( W \) are image dimensions, \( C \) is the number of classes, and \( D \) is the embedding dimension. Our prompt learner constructs class‑specific prompts of the form  
\(\textit{``a photo of a} \texttt{[CLASS]}\textit{."}\),  
where \(\textit{``a photo of a"}\) serves as the initialized learnable context, and \(\texttt{[CLASS]}\) denotes the class name token.  
Classification is performed by computing cosine similarities $\cos(,)$ between the image features $V$ and each class prompt embedding $\boldsymbol{T_p^{(i)}}$, scaling these scores by the temperature $\tau$, applying $\mathrm{Softmax}\bigl(\cos(\boldsymbol{V}, \boldsymbol{T_p^{(i)}})/\tau\bigr)$ to obtain class probabilities $p_i$, and selecting the final label as $\arg\max_i \, p(y=i \mid \boldsymbol{V,T_p^{(i)}})$. 
Following the setup of BiomedCoOp, we generate prompts for each class \( c_i \) in medical modality by querying each of several recent LLMs separately with the standardized instruction:\textit{``Give N textual descriptions of visual discriminative features for distinct medical cases of} \texttt{[CLASS]} \textit{found in} \texttt{[MODALITY]}\textit{."} We treat all LLM-provided template features as samples distributed on a shared manifold space and estimate a class-level semantic prototype field over this space. For each class \( c_i \), we extract a semantic prototype vector \( \boldsymbol{\mu}_{L,c}^{(i)} \), which guides the prompt learner in balancing multi-modal alignment and semantic expressiveness.
% \subsubsection{LLM-Derived Medical Semantic Priors}
\subsection{Inverse Estimation on Hyperspherical Manifold}

In CLIP, both the text and vision encoders produce $\ell_2$-normalized output vectors, placing all textual embeddings $\{ \boldsymbol{w}_i \}$ naturally on the unit hypersphere $S^{d-1}$. Likewise, for each class, $n_p$ prompt templates are generated by each LLM and projected through the CLIP text encoder, yielding prompt embeddings that also lie on $S^{d-1}$. This allows us to formulate a unified embedding space in \( \mathbb{R}^{512} \) by mapping all modalities into a shared hyperspherical manifold, defined as
\begin{equation}
S^{d-1} = \{ \mathbf{x} \in \mathbb{R}^d \mid \|\mathbf{x}\|_2 = 1 \}.
\end{equation}
\subsubsection{Definition 1 (von Mises-Fisher Distribution).} The vMF distribution, which is particularly suited for modeling directional data, defines a probability density over the unit hypersphere \( S^{d-1} \) as:
\(
p_{\mathrm{vMF}}(\mathbf{x}; \boldsymbol{\mu}, \kappa)
= C_d(\kappa)\,\exp\!\bigl(\kappa\,\boldsymbol{\mu}^\top\mathbf{x}\bigr),
\)
where the mean direction \( \boldsymbol{\mu} \in S^{d-1} \) represents the orientation toward which samples concentrate, \( \kappa \geq 0 \) is the concentration parameter controlling the spread around \( \boldsymbol{\mu} \), and \( C_d(\kappa) \) is the normalization constant.

We then fit the vMF distribution to CLIP’s text-token embeddings and LLM-generated prompt embeddings as directional vectors on the hypersphere to map their semantic bias on the hypersphere, which yields a CLIP Semantic Anchor Field $\mathcal{F}_{C}$ and an LLM Semantic Prototype Field $\mathcal{F}_{L,c}$. These two fields are subsequently fused to estimate Unified Semantic Anchors that serve as target attractors to guide prompt optimization on the shared hyperspherical manifold.

\subsubsection{CLIP Semantic Anchor Field.} 
Let \([\boldsymbol{w}_1;\dots;\boldsymbol{w}_V]\in\mathbb{R}^{V\times d}\) denote the vocabulary embedding matrix $\boldsymbol{W}$ from the CLIP text encoder, where \(V\) is the vocabulary size and \(d\) is the embedding dimension, each word embedding \( \boldsymbol{w}_i \in S^{d-1} \) lies on the unit hypersphere. We assume that these embeddings are independent and identically distributed samples drawn from a vMF distribution:
\(
\mathcal{F}_C=p_{\mathrm{vMF}}(\boldsymbol{w}; \boldsymbol{\mu}_C, \hat\kappa_C) = C_d(\hat\kappa_C) \exp(\hat\kappa_C \boldsymbol{\mu}_C^\top \boldsymbol{w}).
\)
We estimate the parameters \( \boldsymbol{\mu}_C \) and \( \hat{\kappa}_C \) using maximum likelihood estimation (MLE), by computing the empirical mean of the vocabulary embeddings \( \bar{\boldsymbol{w}} = \frac{1}{V} \sum_{i=1}^{V} \boldsymbol{w}_i \), the Euclidean norm \( R = \| \bar{\boldsymbol{w}} \|_2 \), and applying:
\begin{equation}
\boldsymbol{\mu}_C = \frac{\bar{\boldsymbol{w}}}{R}, \quad {\kappa}_C \approx \frac{R (d - R^2)}{1 - R^2 + \epsilon}.
\end{equation}
This produces our CLIP Semantic Anchor Field \( \{ {\boldsymbol{\mu}}_C, {\kappa}_C \} \), where \( \epsilon \) is a small constant to avoid division by zero. \( \boldsymbol{\mu}_C \) represents the global semantic direction of CLIP’s vocabulary, and \({\kappa}_C \) quantifies its semantic concentration.

\subsubsection{LLM Semantic Prototype Field.}

For each class \( c \), the LLM generates \( n_p \) prompt templates. Let
\(
\boldsymbol{T}_c = \{ \boldsymbol{t}_{c,1}, \ldots, \boldsymbol{t}_{c,n_p} \} \subset S^{d-1}
\)
denote their CLIP-encoded embeddings. We model these class-specific prompt features using a class-conditional vMF distribution:
\(
\mathcal{F}_{L,c}=p_{\mathrm{vMF}}(\boldsymbol{t}; \boldsymbol{\mu}_{L,c}, \kappa_{L,c}) = C_d(\kappa_{L,c}) \exp(\kappa_{L,c} \boldsymbol{\mu}_{L,c}^\top \boldsymbol{t}).
\)
Similarly, with the empirical mean of class-\( c \) prompt embeddings \( \bar{\boldsymbol{t}}_c = \frac{1}{n_p} \sum_{j=1}^{n_p} \boldsymbol{t}_{c,j} \), and its norm \( R_{L,c} = \| \bar{\boldsymbol{t}}_c \|_2 \), we estimate:
\begin{equation}
\boldsymbol{\mu}_{L,c} = \frac{\bar{\boldsymbol{t}}_c}{R_{L,c}}, \quad 
\kappa_{L,c} \approx \frac{R_{L,c}(d - R_{L,c}^2)}{1 - R_{L,c}^2+ \epsilon}.
\end{equation}
 This defines the LLM Semantic Prototype Field \( (\boldsymbol{\mu}_{L,c}, \kappa_{L,c}) \) for each category \( c \), reflecting the intrinsic semantic prototype bias from the LLM for that class.
\subsubsection{Unified Semantic Anchors.} On the two hyperspherical fields \( \mathcal{F}_C \) and \( \mathcal{F}_{L,c} \), to guide prompt learning, we fuse the global CLIP anchor and the class-specific LLM prototype into a unified semantic anchor \( \boldsymbol{u}_i \) for class \( i \):
\begin{equation}
\boldsymbol{a}_C = \kappa_C \boldsymbol{\mu}_C,\quad
\boldsymbol{c}_i = \kappa_{L,c} \boldsymbol{\mu}_{L,c},\quad
\boldsymbol{u}_i = \frac{\boldsymbol{a}_C + \boldsymbol{c}_i}{\|\boldsymbol{a}_C + \boldsymbol{c}_i\|_2}.
\end{equation}
Each fused vector \( \boldsymbol{u}_i \in S^{d-1} \) serves as a unified semantic anchor, blending CLIP’s global directional prior with class‑specific prototype semantics derived from the LLM, and thereby steering the prompt learner along the shared hyperspherical manifold.

\subsection{Prompt Optimization on Hyperspherical Manifold}
With a unified hyperspherical manifold that integrates global CLIP semantics and class-specific LLM priors, our next step is to learn prompts that are semantically expressive and cross-modally aligned with visual features on this shared hyperspherical space. While conventional prompt learning operates in flat Euclidean space, our method optimizes over a unit hypersphere, where representations lie on a curved Riemannian manifold governed by angular geometry.

Grounded in our task formulation and the hyperspherical space geometry, we considered the prompt optimization to follow three aspects. First, for each class \( c_i \), the prompt embedding \( \boldsymbol{\mathcal{P}}_{c_i} \) is expected to navigate toward its unified semantic anchor \( \boldsymbol{u}_i \), aligning with a balanced directional alignment between CLIP and LLM representations. Second, in few-shot settings with limited supervision, we promote class discriminability by geometrically encouraging prompts to stay angularly distant from irrelevant categories. Finally, prompt optimization should ensure consistency between visual and prompt-based predictions to support cross-modal understanding. The overall optimization is defined as \(
\mathcal{L} = \lambda_\mathit{\boldsymbol{anc}} \, \mathcal{L}_\mathit{\boldsymbol{anc}} + \lambda_{\boldsymbol{sc}} \, \mathcal{L}_\mathit{\boldsymbol{sc}} + \mathcal{L}_\mathit{\boldsymbol{SCE}}
\) with $\lambda_\mathit{\boldsymbol{anc}}$, $\lambda_\mathit{\boldsymbol{sc}}$ as balancing weights.

\subsubsection{Semantic Anchor Loss.}
To align each prompt embedding \( \boldsymbol{\mathcal{P}}_{c_i} \) with its unified semantic anchor \( \boldsymbol{u}_i \),  \textit{Semantic Anchor Loss} explicitly constrains the prompt embedding \( \boldsymbol{\mathcal{P}}_{c_i} \) to approach its unified semantic target \( \boldsymbol{u}_i \) along a directionally controllable path for each class \( {c}_i \) . Instead of altering the anchor itself, we introduce a learnable offset \( \boldsymbol{\delta}_i \in \mathbb{R}^d \) and a global scaling factor \( \alpha \in \mathbb{R}_{>0} \) to dynamically adjust the target direction during optimization. The learnable offset \( \boldsymbol{\delta}_i \) navigates the target direction on the \( {S}^{d-1} \), while \( \alpha \) controls the step size of this deviation. Together, they enable continuous navigation on the hypersphere during optimization:
\begin{equation}
\boldsymbol{u}^d_i = \boldsymbol{u}_i + \alpha \cdot \boldsymbol{\delta}_i, \quad 
\tilde{\boldsymbol{u}}^d_i = \frac{\boldsymbol{u}^d_i}{\|\boldsymbol{u}^d_i\|_2}.
\end{equation}
Here, dynamic target \( \boldsymbol{u}^d_i \) guides the update of the prompt embedding in each training step. The loss is defined as the average squared distance between the normalized prompt embedding \( \tilde{\boldsymbol{\mathcal{P}}_{c_i}} \) and its current directional anchor target \( \tilde{\boldsymbol{u}}^d_i \) across \( C \) classes:
\begin{equation}
\mathcal{L}_\mathit{\boldsymbol{anc}} = \frac{1}{C} \sum_{i=1}^C \left\| \tilde{\boldsymbol{\mathcal{P}}_{c_i}} - \tilde{\boldsymbol{u}}^d_i \right\|_2^2 ,\quad  \tilde{\boldsymbol{\mathcal{P}}}_{c_i} = \frac{\boldsymbol{\mathcal{P}}_{c_i}}{\|\boldsymbol{\mathcal{P}}_{c_i}\|_2}.
\end{equation}
This formulation enables the prompt embedding to adaptively follow class-specific directional guidance over the hyperspherical manifold.

\subsubsection{Spherical Contrastive Loss.}
As shown in \Cref{ov} C, to better address few-shot classification within each class \( c_i \), \textit{Spherical Contrastive Loss} aims to increase the angular separation between the prompt embedding \(\boldsymbol{\mathcal{P}}_{c_i}\) and distractor-class unified anchors \( \boldsymbol{u}_j \), thereby refining the few-shot representations. It does so by formulating two key questions during optimization: 

\textit{1. Why encourage alignment with \( \boldsymbol{u}_i \) rather than \( \boldsymbol{u}_j \)?} On the hypersphere, decision boundaries are defined by equal angular proximity to class anchors. Thus, maximizing the similarity between prompt \( \boldsymbol{\mathcal{P}}_{c_i} \) and its corresponding unified anchor \( \boldsymbol{u}_i \) ensures angular closeness to the correct class. We construct a prototype affinity matrix $\boldsymbol{S}$:
\begin{equation}
\boldsymbol{S} = \tau\,\boldsymbol{P}\,\boldsymbol{U}^\top,\quad S_{ij} = \tau\,\boldsymbol{p}_{c_i}^\top \boldsymbol{u}_j = \tau \cos \theta_{ij},
\end{equation}
where \( \boldsymbol{P} = [\boldsymbol{\mathcal{P}}_1;\ldots;\boldsymbol{\mathcal{P}}_{c_i}] \) and \( \boldsymbol{U} = [\boldsymbol{u}_1;\ldots;\boldsymbol{u}_i] \) are \( C \times d \) matrices of \( \ell_2 \)-normalized prompt features and semantic anchors, respectively, and \( \theta_{ij} \) is the angle between \( \boldsymbol{\mathcal{P}}_i \) and \( \boldsymbol{u}_j \).

\textit{2. How to avoid distractors?} We apply a row-wise softmax cross-entropy across \( C \) classes:
\begin{equation}
\mathcal{L}_\mathit{\boldsymbol{sc}} = -\frac{1}{C} \sum_{i=1}^C \log \frac{\exp(S_{ii})}{\sum_{j=1}^C \exp(S_{ij})},
\end{equation}
which simultaneously pulls each prompt toward its correct anchor and pushes it away from all others. To progressively sharpen angular margins, we anneal the temperature \( \tau \) from \( \tau_0 \) to \( \tau_{\text{max}} \) using a cosine schedule:
\begin{equation}
\tau(t) = \tau_{\text{max}} + \tfrac{1}{2} (\tau_0 - \tau_{\text{max}})\left(1 + \cos\left(\frac{\pi t}{T}\right)\right),
\end{equation}
resulting in original decision boundaries early in training and increasingly discriminative separation as convergence progresses.

\subsubsection{Symmetric Cross-Entropy Loss.}
To tightly couple CLIP’s visual and prompt embeddings, the \textit{Symmetric Cross Entropy Loss}  jointly rewards correct class probability and penalizes any mass on incorrect classes. Given logits 
\(L \in \mathbb{R}^{B \times C}\) produced by the dot product of normalized image features and prompt embeddings, with \(L_{b,c}\) denoting the score of the \(b\)-th sample in the batch \(B\) for the \(c\)-th class among all \(C\) classes, we compute class probabilities \(p_{b,c} = \mathrm{Softmax}(L)_{b,c}\). The loss is defined as:
\begin{equation}
\mathcal{L}_\mathit{\boldsymbol{SCE}} = -\frac{1}{B}\sum_{b=1}^{B}\Bigl[\log p_{b,y_b} + \sum_{c=1}^{C} p_{b,c}\log\bigl(q_{b,c} + \epsilon\bigr)\Bigr],
\end{equation}
where \(y_b\) is the ground-truth label for sample \(b\), and \(q_{b,c} = \mathds{1}_{\{ c = y_b \}}\) is the corresponding one-hot target. The first term encourages confident predictions on the correct class, aligning visual and prompt representations, while the second term penalizes distributional ambiguity by suppressing incorrect class probabilities, with \(\epsilon\) preventing instability in the log-domain. Together, $\mathcal{L}_\mathit{\boldsymbol{SCE}}$ enhances cross-modal alignment by jointly calibrating certainty and discrimination.

% Prompt ensembling is widely used in natural image tasks \cite{zhang2024prefer}, but handcrafted templates show clear limitations in biomedical domains, as they often fail to capture clinical knowledge or provide sufficient semantic diversity \cite{sung2021can}. Clinically informed prompts generated by LLMs have been shown to improve the relevance and semantic diversity of textual descriptions in medical imaging tasks \cite{bhayana2025leveraging}. Different LLMs exhibit substantial variation in their grasp of clinical language, anatomical detail, and medical factuality, leading to inconsistencies in prompt quality \cite{singhal2023large}.
% Prompt ensembling is widely used in natural image tasks, but handcrafted templates show clear limitations in biomedical domains, as they often fail to capture clinical knowledge or offer sufficient semantic diversity. Clinically informed prompts generated by LLMs have been shown to improve the relevance and expressiveness of textual descriptions in medical imaging tasks, but different LLMs exhibit substantial variation in their grasp of clinical language, anatomical detail, and medical factuality, leading to inconsistencies in prompt quality. 

\section{Experiments}
\renewcommand{\arraystretch}{0.9}% reduce row height
\setlength{\tabcolsep}{8pt}% reduce column separation

\begin{table*}[t]
\centering
\scriptsize% further reduce font size for width fit
\caption{Comparison with state-of-the-art methods: average classification accuracy (\%) on 14 biomedical datasets, reported as mean ± standard deviation over 3 random support sets per dataset. Best results are highlighted in bold.}
\label{table:fewshot-main}
\resizebox{1\textwidth}{!}{%
\begin{tabular}{lccccccc}
\toprule
\textbf{Method} & $K=1$ & $K=2$ & $K=4$ & $K=8$ & $K=16$ & $K=32$ & $K=64$ \\
\midrule
% Centered section headings
% \rowcolor{gray!20}
\multicolumn{8}{c}{\hspace{3.8cm}\textbf{Zero-shot Methods}} \\
BiomedCLIP \cite{zhang2023biomedclip} & \multicolumn{7}{c}{\hspace{-2.2cm}$46.32$} \\
BiomedCLIP \cite{zhang2023biomedclip} + Ensemble & \multicolumn{7}{c}{\hspace{-2.2cm}$54.46$} \\
BiomedCLIP \cite{zhang2023biomedclip} + Selective Ensemble & \multicolumn{7}{c}{\hspace{-2.2cm}$56.98$} \\
\midrule
% \rowcolor{gray!20}
\multicolumn{8}{c}{\hspace{4cm}\textbf{CLIP-based Adapter Methods}} \\
CLIP-Adapter \cite{eslami2021does} & $43.52\pm1.25$ & $43.27\pm2.36$ & $45.66\pm2.01$ & $44.98\pm3.13$ & $47.25\pm2.58$ & $48.10\pm1.55$ & $50.85\pm1.40$ \\
Tip-Adapter \cite{zhang2021tip} & $47.58\pm4.36$ & $51.88\pm5.69$ & $59.04\pm4.51$ & $61.37\pm4.62$ & $66.29\pm2.10$ & $67.02\pm2.86$ & $66.45\pm2.97$ \\
Tip-Adapter-F \cite{zhang2021tip} & $49.82\pm7.64$ & $53.34\pm4.36$ & $59.21\pm3.28$ & $60.35\pm2.84$ & $64.27\pm4.33$ & $68.58\pm3.59$ & $68.27\pm0.69$ \\
\midrule
% \rowcolor{gray!20}
\multicolumn{8}{c}{\hspace{4cm}\textbf{Linear Probing Methods}} \\
Standard LP \cite{radford2021learning} & $49.38\pm5.77$ & $52.85\pm5.39$ & $58.78\pm3.87$ & $60.88\pm7.62$ & $65.75\pm4.27$ & $70.80\pm3.45$ & $70.25\pm2.40$ \\
LP++ \cite{huang2024lp++}  & $49.39\pm7.28$ & $58.62\pm5.43$ & $62.39\pm3.10$ & $64.37\pm5.64$ & $69.40\pm2.91$ & $71.50\pm2.30$ & $71.89\pm2.87$ \\
\midrule
% \rowcolor{gray!20}
\multicolumn{8}{c}{\hspace{4.05cm}\textbf{Prompt Learning Methods}} \\
CoOp \cite{zhou2022learning} & $47.63\pm4.52$ & $49.88\pm3.27$ & $54.05\pm2.19$ & $60.32\pm1.07$ & $68.43\pm1.57$ & $72.46\pm1.20$ & $70.30\pm3.25$ \\
CoCoOp  \cite{zhou2022conditional} & $49.43\pm3.25$ & $50.77\pm3.29$ & $54.69\pm4.79$ & $61.08\pm3.49$ & $65.09\pm2.87$ & $66.59\pm6.50$ & $69.75\pm4.10$ \\
KgCoOp \cite{yao2023visual} & $52.46\pm3.89$ & $54.78\pm3.29$ & $58.35\pm4.05$ & $60.78\pm6.34$ & $60.95\pm7.48$ & $71.89\pm3.54$ & $70.36\pm2.38$ \\
ProGrad \cite{zhu2023prompt} & $54.97\pm5.39$ & $56.27\pm4.46$ & $63.28\pm4.23$ & $68.87\pm4.69$ & $70.33\pm4.98$ & $74.39\pm1.58$ & $71.25\pm4.38$ \\
BiomedCoOp \cite{koleilat2025biomedcoop} & $55.08\pm5.85$ & $57.98\pm4.20$ & $63.65\pm3.27$ & $71.29\pm2.19$ & $73.63\pm1.27$ & $75.08\pm3.28$ & $73.65\pm3.98$ \\
\midrule
% Our method row with light gray background
% \rowcolor{gray!7}
\textbf{vMFCoOp (Ours)} & \hspace{0.5mm}$\boldsymbol{57.25 \pm 4.75}$ & \hspace{0.5mm}$\boldsymbol{58.88 \pm 3.95}$ & \hspace{0.5mm}$\boldsymbol{68.29 \pm 2.07}$ & \hspace{0.5mm}$\boldsymbol{72.07 \pm 1.98}$ &\hspace{0.5mm} $\boldsymbol{75.45 \pm 1.48}$ & \hspace{0.5mm}$\boldsymbol{77.08 \pm 1.36}$ & \hspace{0.5mm}$\boldsymbol{77.49 \pm 1.05}$ \\

\bottomrule
\end{tabular}%
}
\end{table*}
We evaluate the vMFCoOp framework under multiple protocols on diverse biomedical benchmarks, including clinically complex datasets like the UK Biobank, to assess accuracy, generalization, and few-shot applicability in realistic medical imaging scenarios. Full details are provided in our \textit{Appendix document}.
\subsection{Experimental Setup}
\subsubsection{Datasets.}We conduct experiments on 14 biomedical imaging datasets spanning 13 anatomical regions and 12 medical imaging modalities, including Cardiac Cine MRI (CMRI), Liver MRI (LMRI) and Pancreas MRI (PMRI) from UK Biobank \cite{sudlow2015uk}; and Endoscopy (Kvasir \cite{pogorelov2017kvasir}), Computerized Tomography (CTKidney \cite{islam2022vision}), Brain MRI (BTMRI \cite{nickparvar2021brain}), Histopathology (LC25000 \cite{borkowski2019lung}, CHMNIST \cite{kather2016multi}), Dermoscopy (DermaMNIST \cite{codella2019skin,tschandl2018ham10000}), Fundus Photography (RETINA \cite{kohler2013automatic}), Optical Coherence Tomography (OCTMNIST \cite{kermany2018identifying}), Ultrasound (BUSI \cite{al2020dataset}), and X-ray (COVID-QU-Ex \cite{tahir2021covid}, KneeXray \cite{chenknee,porwal2018indian}).
For the UK Biobank, we select patients active after July 1st, 2020, and derive fine-grained disease labels using ICD-10 codes \cite{stroganov2022mapping} with descriptions. This diverse and clinically realistic setup enables comprehensive evaluation across varied biomedical imaging conditions. Full details on patient list, data splits, and task definitions are provided in the \textit{Appendix document}.
\subsubsection{Few-Shot Learning.}We follow the few-shot setting of BiomedCoOp to evaluate model performance under limited supervision, using \(K = 1, 2, 4, 8, 16\) labeled samples per class. Additionally, we include evaluations under 32- and 64-shot settings, given that all datasets (except BTMRI) contain over 2,000 training samples. These higher-shot scenarios better reflect realistic clinical conditions and help mitigate overfitting and variance commonly observed in extremely low-shot regimes.
\subsubsection{Base-to-Novel Class Generalization.}To evaluate generalization, we follow BiomedCoOp by splitting each dataset into base and novel classes. Models are trained on base classes with 16-shot supervision and tested on both base and novel classes, assessing their ability to recognize unseen disease types without fine-tuning. Results for 32- and 64-shot settings are included in the \textit{Appendix document}.
\subsubsection{Implementation Details.} For biomedical CLIPs, we adopt BiomedCLIP~\cite{zhang2023biomedclip} with a ViT-B/16 backbone as the default VLM, and additionally evaluate compatibility with PubMedCLIP~\cite{eslami2023pubmedclip}, MedCLIP~\cite{wang2022medclip}, and PMC-CLIP~\cite{lin2023pmc}. We access leading LLMs from major families, including Qwen2.5-72B-Instruct~\cite{qwen2025qwen25technicalreport}, Claude 3.5–1022~\cite{claude2024anthropic}, and DeepSeek R1~\cite{deepseekai2025deepseekr1incentivizingreasoningcapability}. Following BiomedCoOp, we use GPT-4~\cite{openai2024gpt4technicalreport} with 50 prompts per class as the default. To explore the effect of prompt quantity, additional results with 150 prompts per class are provided in the \textit{Appendix document}. We initialize the learnable context using the embedding of \textit{``a photo of a''}. The learning rate is set to 0.003 with a batch size of 4. We use SGD with cosine learning rate scheduling. The constraint weights $\lambda_\mathit{\boldsymbol{anc}}$ and $\lambda_\mathit{\boldsymbol{sc}}$ for each datasets are detailed in the \textit{Appendix document}. All experiments are conducted on an
NVIDIA A100 GPU (80GB).

% \newcolumntype{G}{>{\columncolor{gray!7}}c}
\begin{table}[t]
  \centering
  % \scriptsize
  % \setlength{\tabcolsep}{3pt}        % 列间距
  \renewcommand{\arraystretch}{1.1}   % 行高
  \resizebox{\linewidth}{!}{%
    \begin{tabular}{l c|cccc}
      \toprule
      Dataset            & Metric & BiomedCLIP & CoCoOp    & BiomedCoOp  & \textbf{vMFCoOp} \\
      \midrule
\multirow[c]{3}{*}{\makecell[l]{Avg on\\13 datasets}}
 & Base  & 45.64 & 67.94 & 73.26 & \textbf{78.02} \\
 & Novel & 64.66 & 65.57 & 71.91 & \textbf{76.70} \\
 & \textbf{HM}    & 53.51 & 66.73 & 72.58 & \textbf{77.35} \\

      \midrule
      \multirow[c]{2}{*}{BTMRI} 
                         & Base   & 44.32      & 79.43     & 82.73    & \textbf{83.72} \\
                         & Novel  & 92.57      & 92.64     & 95.26    & \textbf{97.34} \\ \midrule
      \addlinespace[0.8pt]
      \multirow[c]{2}{*}{COVID-QU-Ex} 
                         & Base   & 51.38      & 73.47   & 75.19 &  \textbf{77.42}     \\
                         & Novel  & 89.25      & 89.79     & 91.08    & \textbf{93.09} \\ \midrule
      \addlinespace[0.8pt]
      \multirow[c]{2}{*}{CTKIDNEY} 
                         & Base   & 31.59      & 80.35     & 86.64    & \textbf{87.53} \\
                         & Novel  & 49.21      & 58.42     & 75.09    & \textbf{80.47} \\ \midrule
      \addlinespace[0.8pt]
      \multirow[c]{2}{*}{DermaMNIST} 
                         & Base   & 36.37      & 46.53     & 52.89    & \textbf{55.67} \\
                         & Novel  & 48.55      & 64.85     & 75.28    & \textbf{76.98} \\ \midrule
      \addlinespace[0.8pt]
      \multirow[c]{2}{*}{Kvasir} 
                         & Base   & 73.29      & 85.85     & 86.72    & \textbf{87.35} \\
                         & Novel  & 58.47      & 56.39     & 60.98    & \textbf{62.11} \\ \midrule
      \addlinespace[0.8pt]
      \multirow[c]{2}{*}{CHMNIST} 
                         & Base   & 35.49      & 86.53     & \textbf{88.98}    & 88.35     \\
                         & Novel  & 41.38      & 42.13     & 42.19 & \textbf{53.49} \\ \midrule
      \addlinespace[0.8pt]
      \multirow[c]{2}{*}{LC25000 } 
                         & Base   & 55.42      & 87.98     & 92.91    & \textbf{94.07} \\
                         & Novel  & 86.59      & 94.62     & 96.85    & \textbf{97.80} \\ \midrule
      \addlinespace[0.8pt]
      \multirow[c]{2}{*}{RETINA} 
                         & Base   & 43.52     & \textbf{70.37}     & 68.77    & 70.32     \\
                         & Novel  & 51.89      & 62.29     & 67.30    & \textbf{69.22} \\ \midrule
      \addlinespace[0.8pt]
      \multirow[c]{2}{*}{KneeXray} 
                         & Base   & 31.07      & 29.85     & 42.48    & \textbf{50.19} \\
                         & Novel  & 68.43      & 62.59     & 79.32    & \textbf{79.05} \\ \midrule
      \addlinespace[0.8pt]
      \multirow[c]{2}{*}{OCTMNIST} 
                         & Base   & 53.25      & 79.54     & 79.32    & \textbf{80.08} \\
                         & Novel  & 49.57      & \textbf{51.37} & 49.27 & 51.25     \\ \midrule
      \addlinespace[0.8pt]
      \multirow[c]{2}{*}{CardiacMRI} 
                         & Base   & 48.25        & 53.46        & 52.59       & \textbf{64.59}         \\
                         & Novel  & 52.78         & 49.78        & 47.36     & \textbf{70.85}         \\ \midrule
      % \addlinespace[0.8pt]
      % \multirow[c]{2}{*}{CMRI-LAX} 
      %                    & Base   & --         & --        & --       & --         \\
      %                    & Novel  & --         & --        & --       & --         \\
      \addlinespace[0.8pt]
      \multirow[c]{2}{*}{LiverMRI}  
                         & Base   & 43.77         & 39.29        & 48.29       & \textbf{66.39}        \\
                         & Novel  & 64.38         & 48.98        & 66.59       &  \textbf{73.97}        \\ \midrule
      \addlinespace[0.8pt]
      \multirow[c]{2}{*}{PancreasMRI}  
                         & Base   & 45.62         & 70.59        & 74.89       &  \textbf{83.58}        \\
                         & Novel  & 87.55         &  78.56        & 88.23       &  \textbf{91.45}        \\
      \bottomrule
    \end{tabular}%
 }
  \caption{Base-to-Novel Generalization: Accuracy (\%) Comparison of BiomedCoOp and SOTA Prompting Approaches.}
  \label{tab:base-to-new-final}
\end{table}

\subsection{Few-shot Benchmarking}
We benchmark our method against five representative text-only prompt learning baselines (CoOp, CoCoOp, KgCoOp, ProGrad, BiomedCoOp), three CLIP-based adapter methods (CLIP-Adapter, Tip-Adapter, Tip-Adapter-F), and two linear probing approaches (Standard LP, LP++), all methods that use BiomedCLIP as the backbone are tuned to their optimal settings. Following BiomedCoOp, we focus on shallow text-side adaptations, excluding joint text-image prompt tuning methods (e.g., MaPLe) to ensure fair comparison. As shown in \Cref{table:fewshot-main}, vMFCoOp consistently outperforms all baselines, with a notable relative gain of 7.29\% at 4-shot and 5.22\% at 64-shot over the second-best method, BiomedCoOp. These two settings are particularly reflective of clinical conditions, where either only a few labeled cases are available or moderate supervision is feasible. Beyond raw accuracy, we observe that while linear probes benefit from more data, their performance quickly saturates. Prompt learning methods improve with more shots but exhibit diminishing variance reduction, and in some cases (e.g., CoOp), performance degrades due to overfitting. In contrast to competing approaches whose gains plateau or even reverse at higher shot counts, vMFCoOp sustains both accuracy improvements and low variance across all $\mathit{K}$-shot regimes. This robustness derives from our unified hyperspherical optimization, which preserves the intrinsic semantic topology between LLM and CLIP embeddings and curbs semantic drift.

\subsection{Base-to-Novel Generalization}
We evaluate the generalization capability of vMFCoOp against state-of-the-art prompt-learning methods from base to novel biomedical categories using harmonic mean (HM) accuracy to balance performance across both sets. Due to limited class diversity, the BUSI dataset is excluded from this experiment. Results presented in \Cref{tab:base-to-new-final} demonstrate that vMFCoOp substantially enhances generalization, achieving an average HM of 77.35\%, surpassing BiomedCoOp by approximately 4.8\%. Notably, vMFCoOp consistently outperforms baseline methods across datasets, particularly in clinically nuanced cases such as BTMRI, CTKIDNEY, and cardiac MRI, reflecting improvements in both base-class and novel-class recognition. This substantial gain is attributed to vMFCoOp’s ability to mitigate semantic misalignment through the unified hyperspherical manifold, enabling more precise multimodal semantic alignment, thus providing robust generalization across diverse and complex clinical imaging conditions.

\subsection{Visual Interpretability}
\begin{figure*}
    \centering
    
    \includegraphics[width=1\textwidth]{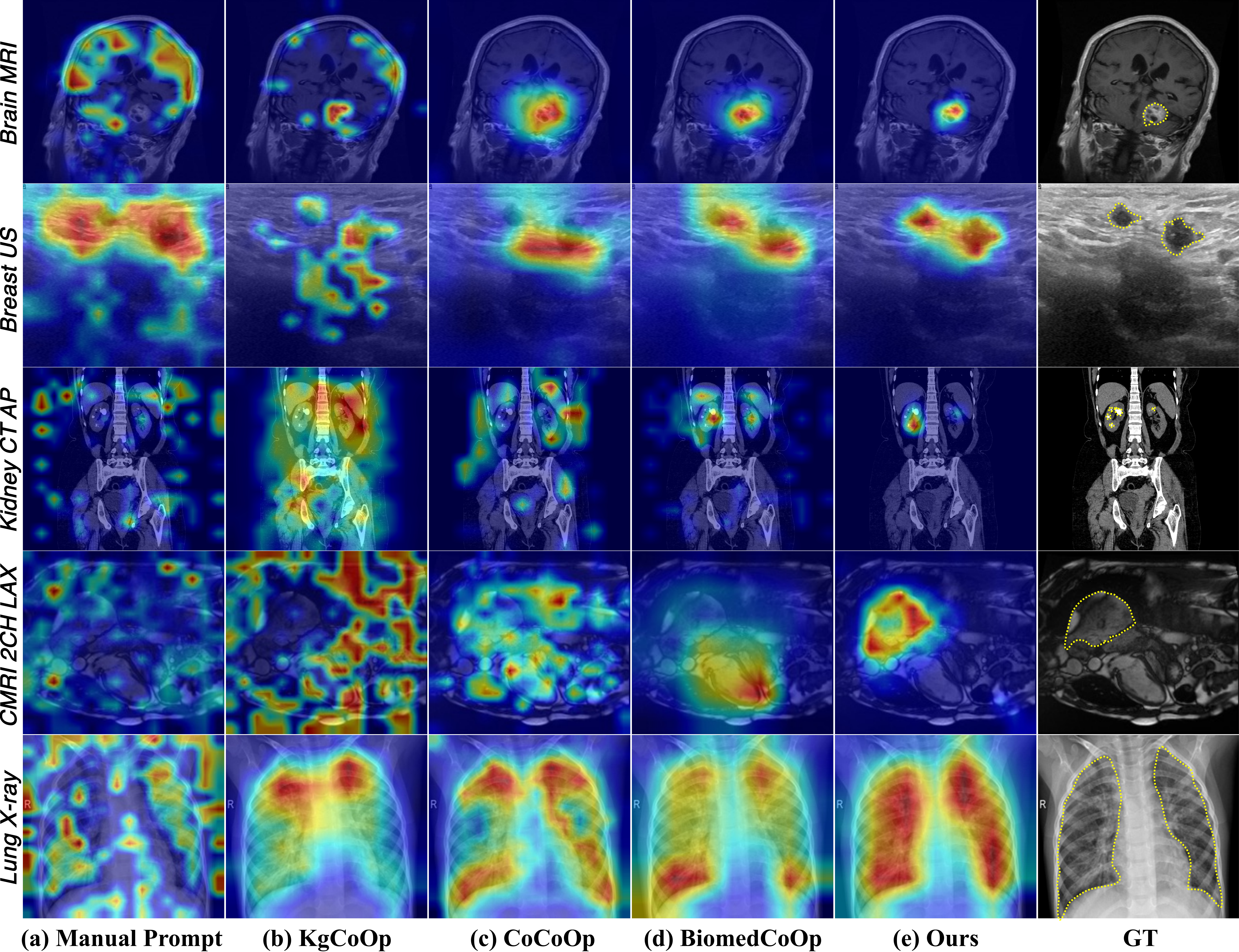}
  \vspace{20pt}   
\caption{Effect of prompting variations on saliency maps, where (a)–(e) illustrate different strategies (zoom in for details). 
Note that the fourth row depicts a rare cardiac cine MRI case in which the patient has a posterior mediastinal tumor, 
simulating a few-shot fine-tuning scenario with limited data and reflecting real clinical settings. 
vMFCoOp (ours) successfully localizes the approximate lesion region under such challenging conditions, while other methods, such as BiomedCoOp, tend to focus their attention on the cardiac area. This may occur because they fail to capture the concept of the posterior mediastinum when classifying \textit{“Malignant neoplasm of heart, mediastinum, and pleura”}, or because their attention is overly biased toward the heart region due to semantic inductive bias, leading to overfitting or semantic misalignment.}

   \label{cp}
     
\end{figure*}
Interpretability is essential in clinical decision-making. Follow BiomedCoOp, we adopt gScoreCAM \cite{Chen_2022_ACCV} to generate saliency maps for five representative samples spanning diverse modalities. As shown in \Cref{cp}, ground-truth lesion contours are overlaid in yellow on the raw images for reference. (a) Manual Prompt with only \(\textit{``a photo of a} \texttt{[CLASS]}\textit{."}\) yields scattered focus and often misses fine details. (b)  KgCoOp adds manual priors, improving low-level localization (e.g., lung fields), but remains diffuse in complex cases (e.g., tiny kidney stones). (c) CoCoOp learns prompts without prior knowledge, leading to highly unstable and noisy saliency across all examples. (d) BiomedCoOp, with LLM-derived textual priors, better captures coarse organ-level semantics but exhibits modality-dependent bias, it focuses narrowly on the heart region in CMRI and misses the obvious posterior mediastinal tumor. In contrast, our (e) vMFCoOp consistently highlights lesion-centric regions aligned with ground truth, demonstrating superior interpretability and robust cross-modal grounding.
% 单栏、紧凑型表格，使用resizebox控制整体大小
\begin{table*}[t]
  \centering
  \renewcommand{\arraystretch}{1}  % 缩减行高
  \caption{Ablation of vMFCoOp components with 3 constraints on few-shot and base-to-novel accuracy (\%).}% 缩减列间距
  \resizebox{0.6\linewidth}{!}{%
  \begin{tabular}{ccc ccc ccccc}
    \toprule
    \multicolumn{3}{c}{\textbf{Components}} & \multicolumn{3}{c}{\textbf{Base-to-Novel}} & \multicolumn{5}{c}{\textbf{Few-shot}} \\
    \cmidrule(lr){1-3} \cmidrule(lr){4-6} \cmidrule(lr){7-11}
    \textbf{$\mathcal{L}_{\boldsymbol{SCE}}$} & \textbf{$\mathcal{L}_{\mathit{\boldsymbol{anc}}}$} & \textbf{$\mathcal{L}_{\mathit{\boldsymbol{sc}}}$} & \textbf{Base} & \textbf{Novel} & \textbf{HM} & \textbf{1} & \textbf{4} & \textbf{8} & \textbf{16} & \textbf{32} \\
    \midrule
    \xmark & \xmark & \xmark & 68.43 & 42.11 & 52.14 & 43.22 & 47.81 & 55.27 & 60.90 & 62.34 \\
    \xmark & \cmark & \xmark & 72.65 & 69.38 & 70.98 & 50.29 & 53.68 & 60.29 & 63.87 & 70.51 \\
    \xmark & \cmark & \cmark & 73.26 & 73.59 & 73.42 & 49.87 & 54.23 & 60.58 & 73.25 & 74.02 \\
    \cmark & \xmark & \xmark & 69.20 & 47.58 & 56.39 & 46.35 & 48.98 & 57.34 & 65.72 & 67.77 \\
    \cmark & \cmark & \xmark & 74.35 & 74.08 & 74.71 & 54.88 & 63.20 & 70.53 & 73.58 & 75.34 \\
    \cmark & \cmark & \cmark & \textbf{78.11} & \textbf{76.22} & \textbf{77.15} & \textbf{57.25} & \textbf{68.29} & \textbf{72.07} & \textbf{75.45} & \textbf{77.08} \\
    \bottomrule
  \end{tabular}%
  }
  
  \label{tab:ablation-components}
\end{table*}

\subsection{Ablation Study}

\subsubsection{Effect of Main Components.}
We summarize the three hierarchical constraints that define the core components of vMFCoOp. In this ablation, we systematically isolate each component under identical settings: \textbf{$\mathcal{L}_{\mathit{\boldsymbol{SCE}}}$} denotes replacing standard cross-entropy with Symmetric Cross-Entropy Loss atop BiomedCLIP; \textbf{$\mathcal{L}_{\mathit{\boldsymbol{anc}}}$} introduces a unified semantic anchor field via inverse vMF estimation on the hyperspherical manifold; and \textbf{$\mathcal{L}_{\mathit{\boldsymbol{sc}}}$} adds Spherical Contrastive Loss to enforce angular margins. As shown in Table~\ref{tab:ablation-components}, each constraint provides incremental improvements, while their combination yields the strongest gains. Notably, incorporating \textbf{$\mathcal{L}_{\mathit{\boldsymbol{anc}}}$} consistently enhances both few-shot performance and base-to-novel generalization, underscoring the importance of aligning multimodal representations via a shared semantic anchor. Adding \textbf{$\mathcal{L}_{\mathit{\boldsymbol{sc}}}$} further improves novel class accuracy, especially in higher-shot regimes, by reducing overfitting through explicit angular separation. While \textbf{$\mathcal{L}_{\mathit{\boldsymbol{SCE}}}$} alone offers modest improvements, its integration with the other two constraints amplifies robustness and cross-modal alignment. These results collectively demonstrate that the three constraints are complementary and jointly contribute to optimal generalization and stability in biomedical few-shot learning. Please note that the base-to-novel results in \Cref{tab:ablation-components} slightly differ from those presented in \Cref{tab:base-to-new-final} because the BUSI dataset was excluded in \Cref{tab:base-to-new-final}, as mentioned earlier.
\subsubsection{Effect of Backbone Configurations.}
\begin{figure}[t]
    \centering
    
    \includegraphics[width=1\linewidth]{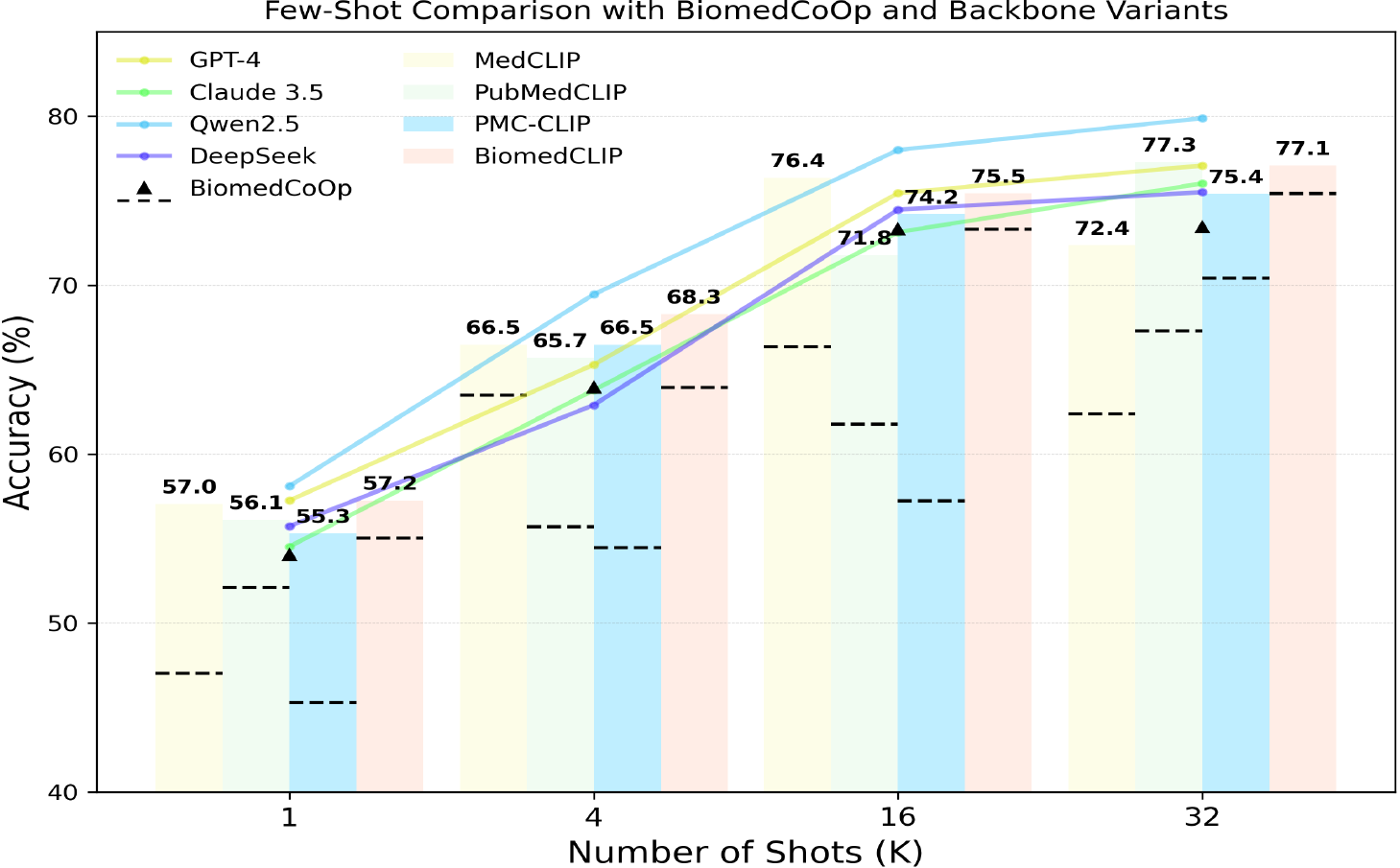}
    
  \caption{Few-shot performance comparison under different backbone configurations using 50 LLM-derived prompts. Black dashed line and triangle denote BiomedCoOp.}
  \label{bv}
\end{figure}

To further validate vMFCoOp’s plug-and-play flexibility across different foundation backbones, \Cref{bv} contrasts performance under two complementary views. The vertical bars show few-shot accuracy for various CLIP variants when paired with GPT-4, with the black dashed line indicating the corresponding BiomedCoOp baseline. Conversely, the overlaid curves trace accuracy across four different LLMs using BiomedCLIP as the fixed visual backbone, where black triangles mark the average performance over all LLMs. Notably, vMFCoOp consistently outperforms BiomedCoOp across all backbone combinations, yielding both higher accuracy and a markedly flatter response curve, demonstrating robust, stable gains regardless of the underlying CLIP or LLM choice. \textit{Appendix document} provides detailed results, including analysis of how the number of LLM prompts affects performance.

\section{Conclusion}
Our work bridges the critical semantic gap in prompting biomedical vision-language models by proposing a unified geometric framework that elegantly aligns heterogeneous textual biases from diverse foundational models. By moving beyond conventional Euclidean methods and leveraging the structured semantic manifold defined by vMF distributions, our approach significantly advances few-shot adaptation in complex biomedical imaging scenarios. This unified geometric perspective not only resolves multimodal semantic misalignment but also offers a scalable and principled solution for cross-modal integration in evolving clinical AI.

\section{Future Work}
With the rapid evolution of deep learning \cite{cao2024domain,cao2024domain2,li2025unified,li2025bp,chen2024hint,zhang2024depth,sun2025gl,sun2025bs,chen2024bs,qiu2025adaptively,zhai2024challenges,deng20213d} and multimodal foundation models, future studies will extend this work to a wider range of downstream tasks and natural image domains to further assess its scalability and generalization. We also plan to expand this study into an extended journal version, providing deeper theoretical insights, comprehensive experiments, and broader clinical and natural-domain evaluations.

\section{Ethics Statement}
The UK Biobank received ethical approval from the NHS National Research Ethics Service North West (11/NW/0382; 16/NW/0274), and no additional ethics review was required for this study.

\section{Acknowledgments}
This research was conducted using data from the UK Biobank Resource under Application Number 603483, as part of an existing approved project. The authors express their sincere gratitude to all participants and research personnel involved in the UK Biobank initiative. This work was further supported by the Scheme Towards Collaborative Cloud-Edge Deep Learning Deployment (Grant IEC/NSFC/223523), the National Edge AI Hub for Real Data: Edge Intelligence for Cyber-Disturbances and Data Quality (EP/Y028813/1), and the UK Medical Research Council (MRC) Innovation Fellowship (Grant MR/S003916/2).
 % Interpretability is essential in clinical decision-making. Follow BiomedCoOp, we adopt gScoreCAM to generate saliency maps for five representative samples spanning diverse modalities. As shown in \Cref{cp}, ground-truth lesion contours are overlaid in yellow on the raw images for reference. Columns (a)–(e) illustrate the effect of five prompting strategies: (a) Manual Prompt uses only the basic template \(\textit{`a photo of a} \texttt{[CLASS]}\textit{."}\) resulting in dispersed attention that often fails to localize fine-grained structures, such as the tumor boundary in Brain MRI or micro-calcifications in Kidney Abdomen-Pelvic CT. (b) KgCoOp incorporates manual prompts as prior knowledge, improving localization of low-level features like organ boundaries (e.g., lung fields in X-ray), yet still suffers from diffuse focus under complex anatomy. (c) CoCoOp learns prompts without prior knowledge, leading to highly unstable and noisy saliency across all examples. (d) BiomedCoOp, with LLM-derived textual priors, better captures coarse organ-level semantics but exhibits modality-dependent bias—for example, it focuses narrowly on the heart region in Cardiac MRI and misses the posterior mediastinal lesion. In contrast, our (e) vMFCoOp consistently produces sharp, lesion-aligned saliency maps that closely match ground-truth contours across all modalities. These observations validate the superior interpretability and cross-modal grounding of our method in few-shot biomedical settings.

\bibliography{aaai2026}

\end{document}